\documentclass[runningheads, orivec]{llncs}
\usepackage[T1]{fontenc}
\usepackage{graphicx}
\usepackage{svg}
\usepackage{amsmath}
\usepackage{amssymb}
\usepackage[linesnumbered,ruled,vlined]{algorithm2e}
\begin{document}
\title{SCOPE for Hexapod Gait Generation}
%
%
\author{Jim O'Connor\orcidID{0009-0008-9917-5682} \and
Jay B. Nash\orcidID{0009-0006-3979-2043} \and \\ 
Derin Gezgin\orcidID{0009-0004-0707-603X} \and
Gary B. Parker\orcidID{0009-0001-3870-1190}}
\authorrunning{O'Connor et al.}
%
\institute{Connecticut College, New London CT 06320-4125, USA \\
\email{\{joconno2,jnash1,dgezgin,parker\}@conncoll.edu}}
\maketitle              
\begin{abstract}
Evolutionary methods have previously been shown to be an effective learning method for walking gaits on hexapod robots. However, the ability of these algorithms to evolve an effective policy rapidly degrades as the input space becomes more complex. This degradation is due to the exponential growth of the solution space, resulting from an increasing parameter count to handle a more complex input. In order to address this challenge, we introduce Sparse Cosine Optimized Policy Evolution (SCOPE). SCOPE utilizes the Discrete Cosine Transform (DCT) to learn directly from the feature coefficients of an input matrix. By truncating the coefficient matrix returned by the DCT, we can reduce the dimensionality of an input while retaining the highest energy features of the original input. We demonstrate the effectiveness of this method by using SCOPE to learn the gait of a hexapod robot. The hexapod controller is given a matrix input containing time-series information of previous poses, which are then transformed to gait parameters by an evolved policy. In this task, the addition of SCOPE to a reference algorithm achieves a 20\% increase in efficacy. SCOPE achieves this result by reducing the total input size of the time-series pose data from 2700 to 54, a 98\% decrease. Additionally, SCOPE is capable of compressing an input to any output shape, provided that each output dimension is no greater than the corresponding input dimension. This paper demonstrates that SCOPE is capable of significantly compressing the size of an input to an evolved controller, resulting in a statistically significant gain in efficacy.

\end{abstract}
\section{Introduction} 
\label{sec:introduction}

Gait generation on a hexapod robot has been extensively explored within the field of evolutionary robotics, and is known to be a difficult challenge for learning algorithms to approach \cite{evorobots}. In many works exploring this problem, evolutionary computation has emerged as the natural choice of algorithm \cite{parkergait,adaptive,parkerinciment,lewis}. Initial efforts to learn effective walking gaits on hexapod robots employed a variety of techniques, including genetic programming to learn gait parameters \cite{Busch2002}, genetic algorithms to optimize the parameters of a controlling neural network \cite{Gallagher1996}, and task-specific approaches such as the cyclic genetic algorithm \cite{Parker2000EvolvingHG}. More recent efforts have found success even as the hexapod robot traverses uneven terrain or sustains damage \cite{Vice2022,damage}.

Many of these works focus purely on developing a steady walking gait, with few sensor inputs. For example, \cite{Angliss2023CoevolvingHL} uses a genetic algorithm to evolve the weights of a distributed neural network. However, this network only accepts a total of 18 sensor inputs, 3 for each leg. This limited input size rules out more complex sensor inputs or temporal information. We present SCOPE as a solution to this issue, demonstrating that it can learn an effective walking gait from 2,700 input parameters. This increase in input dimensionality enables the use of temporal and spatial information from the robot’s previous poses, which would otherwise be infeasible for traditional methods. By leveraging this expanded input, SCOPE allows the policy to condition its gait generation on a more informative state representation.

A straightforward response to the increased input dimensionality would be to proportionally increase the number of free parameters in the policy. However, this approach quickly becomes inefficient, as not all input features contribute equally to the control task. Some components of the input may be redundant or irrelevant. For instance, the precise position of each joint at every timestep may be unnecessary if the total displacement over time suffices. Conversely, features such as joint acceleration at each timestep may be critical if they convey information not captured by simpler statistics.

In order to avoid exploding the number of free parameters in the policy, we propose a lightweight, domain-agnostic compression method based on the Discrete Cosine Transform (DCT) \cite{gilbert,dctbook}. The DCT is a well explored tool in signal processing and compression, widely used in applications such as image encoding \cite{DCT,watson}, where it is valued for its ability to concentrate signal energy into a small number of low-frequency components \cite{RAO}.

When applied to neural learning methods, the DCT has been shown to be capable of both a type of pseudo-attention and augmenting standard attention methods \cite{Scribano2023,dang,moredct}. The DCT has been applied to extract additional features from input images to classifier models, which resulted in an increase classification accuracy on standard benchmarks \cite{lee}. We aim to leverage this pseudo-attention as an all-in-one dimensionality reduction solution, enabling an evolved policy to process a much larger number of inputs than otherwise would be feasible.

In our approach, we apply a two-dimensional DCT to the raw input matrix that represents the recent temporal state of the robot. We then truncate the DCT coefficient matrix by retaining a fixed-size top-left block corresponding to the lowest-frequency components in both temporal and spatial dimensions. This fixed truncation reduces dimensionality while preserving the most structurally significant features of the input data, discarding only higher-frequency variations that typically encode noise or fine-grained detail.
 
This compressed representation serves as the input to a compact policy model, evolved using a steady-state genetic algorithm. Despite operating in a drastically reduced space, the controller is able to generate effective and adaptive gait strategies. As we show, this approach provides a favorable tradeoff between expressiveness and tractability. It allows the evolutionary process to scale to high-dimensional inputs without overwhelming computational cost.

\section{Discrete Cosine Transform}
\label{sec:DCT}

The Discrete Cosine Transform (DCT) is a widely used signal transformation technique that expresses a finite sequence of real numbers as a sum of cosine functions oscillating at different frequencies. Unlike the Fourier transform, which uses complex exponentials, the DCT relies solely on real-valued basis functions, making it particularly attractive for real-world applications such as image compression, audio processing, and learning systems. 

Among the various types of DCT, the type-II DCT is the most frequently applied and the one applied in this paper. It is especially known for its energy compaction property: for many natural signals, a large portion of the total energy is concentrated in just a few low-frequency coefficients. This property makes the type-II DCT ideal for dimensionality reduction, as high-frequency coefficients (which often correspond to noise or minor detail) can be discarded without significant information loss.

Formally, for a 1D input vector \( \mathbf{x} = (x_0, x_1, \dots, x_{N-1}) \in \mathbb{R}^N \), the type-II DCT produces an output vector \( \mathbf{X} = (X_0, X_1, \dots, X_{N-1}) \) given by:
\begin{equation*}
X_k = \alpha_k \sum_{n=0}^{N-1} x_n \cos\left[ \frac{\pi}{N} \left(n + \frac{1}{2} \right) k \right], \quad \text{for } k = 0, 1, \dots, N-1,
\end{equation*}
where
\begin{equation*}
\alpha_k = 
\begin{cases}
\sqrt{\frac{1}{N}}, & \text{if } k = 0, \\
\sqrt{\frac{2}{N}}, & \text{otherwise}.
\end{cases}
\end{equation*}
This normalization ensures orthogonality and energy preservation under the transform.

SCOPE is based on the 2D DCT, which extends this definition to matrices. Given a real matrix \( \mathbf{M} \in \mathbb{R}^{m \times n} \), its 2D type-II DCT is computed by applying the 1D DCT to each row and then to each column (or vice versa, due to separability). The result is a matrix \( \mathbf{C} \in \mathbb{R}^{m \times n} \) of DCT coefficients:
\begin{equation*}
\label{DCTeq}
\mathbf{C} = \mathcal{D}_2(\mathbf{M}) = \mathbf{A}_m \mathbf{M} \mathbf{A}_n^\top,
\end{equation*}
where \( \mathbf{A}_m \in \mathbb{R}^{m \times m} \) and \( \mathbf{A}_n \in \mathbb{R}^{n \times n} \) are DCT basis matrices whose rows contain the type-II DCT basis vectors for lengths \( m \) and \( n \), respectively. $\mathcal{D}_2(\cdot)$ denotes the two-dimensional DCT, applied first along the columns, then along the rows.

This decomposition enables a compact representation of \( \mathbf{M} \), especially when truncated to retain only the low-frequency components in the top-left submatrix of \( \mathbf{C} \). In many practical contexts, such truncation retains the dominant structural information while eliminating higher-frequency variations that are less critical for downstream processing.

We utilize the 2D type-II DCT as a domain-agnostic tool for feature extraction and dimensionality reduction, which forms the foundation of the input compression scheme proposed by SCOPE.

\section{Methodology}
\label{sec:Methodology}

\begin{figure*}[t]
    \centering
    \includegraphics[width=1\linewidth]{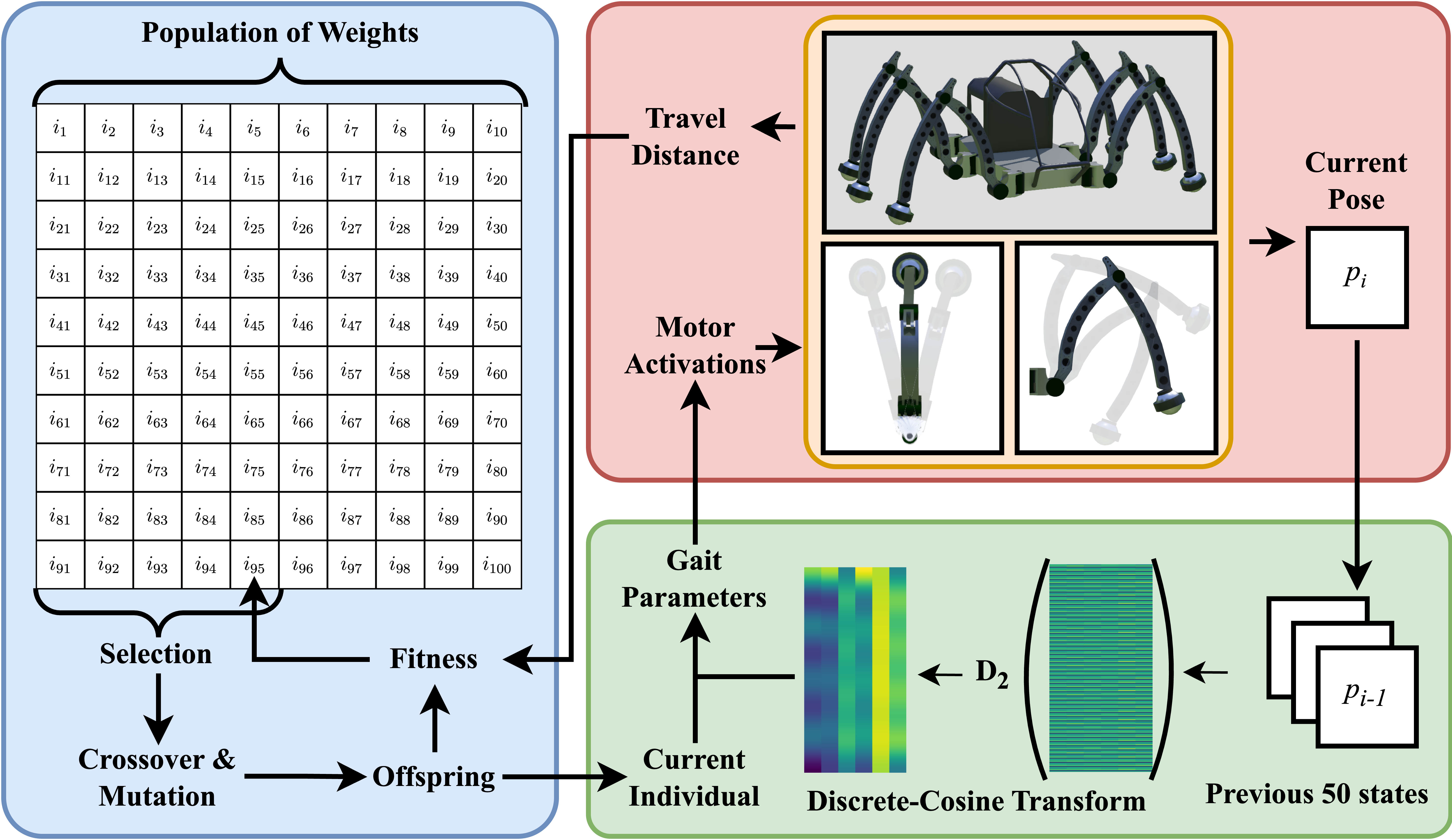}
    \caption{Full methodology overview. The Steady-State Genetic Algorithm provides evolved policy weights for SCOPE, which decides the gait generation parameters for the Mantis robot based on the previous 50 states. A solution is assigned the distanced traveled as its fitness, and is used to replace the worst individual in the tournament selection.}
    \label{fig:main}
\end{figure*}

\subsection{Sparse Cosine Optimized Policy Evolution}
\label{subsec:SCOPE}

Sparse Cosine Optimized Policy Evolution (SCOPE) is an evolutionary algorithm that focuses on high-dimensional control problems, aiming to use dimensionality reduction techniques to allow an evolutionary approach to problems with large input spaces. The properties of the Discrete Cosine Transform (DCT) are leveraged to generate a compressed representation of the input that retains the maximum amount of information, while reducing the number of required parameters. Specifically, SCOPE utilizes the type-II DCT. While there exists eight variants of the DCT, along with many related transforms, the most commonly used variant is the type-II DCT.

During evaluation, the controller receives a high dimensional input which is transformed into output gait parameters. In the method presented in this paper, the input includes the last 50 poses of the robot. These poses consist of the position, velocity and acceleration values for each motor which form a total input size of $2700$ values. While directly optimizing the weights to transform this input into gait parameters is an option, this would require an extremely large chromosome leading to a high computational load. 

\begin{figure}[t]
\centering

\newcommand{\rowheight}{1cm}
\newcommand{\labelW}{0.3\linewidth}
\newcommand{\imagewidth}{0.7\linewidth}

\newcommand{\dctrow}[2]{%
\begin{minipage}[c][\rowheight][c]{\linewidth}
    \begin{minipage}[c]{\labelW}
        \centering #1
    \end{minipage}%
    \begin{minipage}[c]{\imagewidth}
        \includegraphics[width=\linewidth]{#2}
    \end{minipage}
\end{minipage}\vspace{0.3em}
}

\dctrow{Original Pose (6×450)}{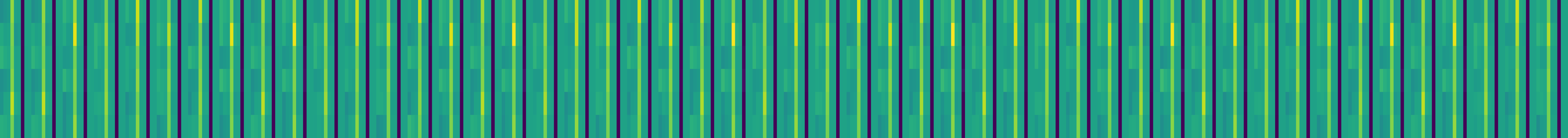}
\dctrow{DCT truncated to 6×54}{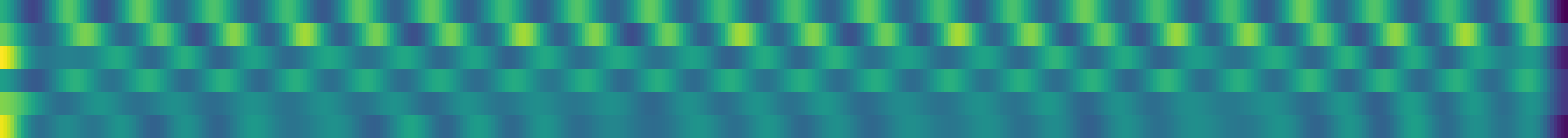}
\dctrow{DCT truncated to 6×9}{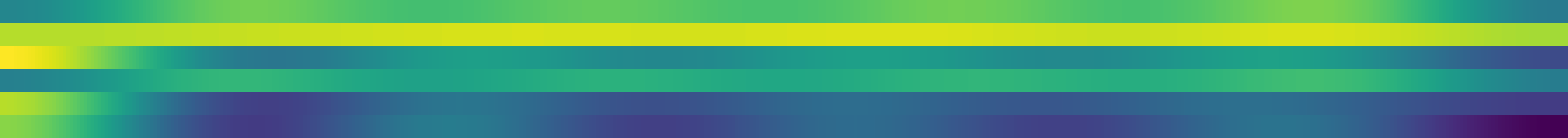}

\caption{Comparison of raw and truncated inputs visualized as heatmaps. The original matrix ($6 \times 450$) is compressed using the Discrete Cosine Transform to $6 \times 54$ and $6 \times 9$, demonstrating preservation of low-frequency features with strong compression.}
\label{fig:dct_compression_visual}
\end{figure}

To address this, SCOPE applies a 2D DCT to the input matrix $\mathbf{M} \in \mathbb{R}^{m \times n}$, where $m = 6$ represents the number of legs, and $n = 450$ represents the concatenation of $50$ consecutive input frames, each with $9$ features, 3 per motor. These frames are placed side by side, resulting in a matrix with 6 rows and $9 \times 50 = 450$ columns. A 2D DCT is applied to this matrix to obtain a frequency-domain representation:
\[
\mathbf{C} = \mathcal{D}_2(\mathbf{M}),
\]
where $\mathcal{D}_2(\cdot)$ denotes the two-dimensional DCT, applied first along the columns, then along the rows. The resulting matrix $\mathbf{C} \in \mathbb{R}^{6 \times 450}$ contains DCT coefficients ordered by frequency. To reduce dimensionality, we retain only the top-left $k_1 \times k_2$ block of $\mathbf{C}$, which contains the lowest-frequency components:
\[
\mathbf{C'} = \mathbf{C}_{1:k_1,\, 1:k_2},
\]
where $\mathbf{C'} \in \mathbb{R}^{k_1 \times k_2}$. In our experiments, we use $k_1 = 6$ and $k_2 = 9$, reducing the original input size from $2700$ to $54$ values, a compression rate of 98\%. The impact of different truncation levels on the DCT-transformed input matrix is visualized in Figure~\ref{fig:dct_compression_visual}, demonstrating how low-frequency components preserve the most relevant structure even under strong compression. This matrix $\mathbf{C'}$ could be further modified by explicitly zeroing out the lowest $p$-percentile of coefficients by absolute magnitude. This process would produce a sparse version of $\mathbf{C'}$, potentially eliminating irrelevant signal components. However, this form of sparsification is only advantageous in the presence of noise or redundancy within the input. In our case, the input matrix encodes precise time-series pose information of a simulated robot, with no stochastic corruption or external noise sources. As a result, further sparsification would offer limited benefit and could instead lead to information loss. For this reason, we refrain from applying percentile-based sparsification in this context.

The compressed matrix $\mathbf{C'}$ is then flattened and passed to a policy evolved by a Steady-State Genetic Algorithm (Section~\ref{subsec:SSGA}). The policy is parameterized by an evolved weight vector $\vec{w}$ and bias vector $\vec{b}$, and produces an output vector $\vec{o}$ via:
\[
\vec{o} = \vec{w} \cdot \operatorname{vec}(\mathbf{C'}) + \vec{b},
\]
where $\operatorname{vec}(\mathbf{C'})$ denotes the column-wise vectorization of $\mathbf{C'}$. The result $\vec{o}$ defines the gait parameters used to actuate the robot.

\subsection{Converting Gait Parameters to Motor Actuation}
\label{subsec:sinegait}

To actuate the motors of the hexapod robot, we use a sinusoidal central pattern generator (CPG) to control each motor. The output vector of the evolutionary policy is split into groups of three values per motor: phase $\phi$, amplitude $A$, and offset $\mu$. These are used to calculate the motor target positions as sinusoidal functions:  

\[
\theta_i(t) = \mu_i + A_i \cdot \sin(2\pi t + \phi_i)
\]

For each timestep, the controller computes the target position $\theta_i(t)$ which is then constrained to obey a set $\Delta\theta$ to ensure smooth transition between positions and prevent unrealistic motor speeds. Moreover, to avoid unrealistic jumps in the motor positions, the total positional change of the motor is split into multiple time-steps via linear slicing from $\theta_i(t-1)$ to $\theta_i(t)$.

This continuous loop allows smooth movement while being simple enough to evolve with a relatively small number of parameters. The gait policy is updated every 15 seconds with newly evolved parameters.

\subsection{Steady State Genetic Algorithm}
\label{subsec:SSGA}

To optimize the parameters of SCOPE, we make use of a Steady State Genetic Algorithm (SSGA). Unlike a standard genetic algorithm which replaces the entire population at each iteration, an SSGA performs incremental updates by replacing a few individuals at each generation. This allows the algorithm to adapt to its environment and act as a real time learning algorithm. In our implementation, the controlling individual is replaced on the robot for each evaluation, but the overall simulation is never paused or reset.

The SSGA is initialized with 100 randomly generated individuals. At the beginning of the training, each individual is tested and their fitness scores are recorded. Once each individual is assigned a starting fitness, we use standard tournament selection to select individuals for crossover and mutation. In each generation, a subset of five individuals is sampled from the overall population and the two individuals with the highest fitness scores are selected for reproduction. Single point crossover is applied to produce two offspring, which are then subjected to Gaussian mutation with a mutation rate of $0.003$ and a standard deviation ($\sigma$) of $0.5$.

These offspring are evaluated sequentially onboard the robot. The euclidean distance traveled by the robot in the simulation while a given offspring is acting as the controller is the fitness score of that offspring. No other information is used to create or modify the fitness score; individuals are free to choose any strategy to move the robot as far as possible. The two lowest-performing individuals in the previously selected tournament are then replaced by the offspring in the overall population. This process is repeated for a fixed number of generations. Throughout evolution, the algorithm maintains and updates the best individual found so far. The step-by-step training and evolution loop is outlined in Algorithm~\ref{alg:ssga}.

\begin{algorithm}[!ht]
\DontPrintSemicolon
\caption{Steady-State Genetic Algorithm}
\label{alg:ssga}

\KwIn{ \\ 
Population size $N$ \\
chromosome length $L$ \\
mutation rate $\mu$ \\
mutation scale $\sigma$ \\ 
max generations $G$}
\KwOut{Best individual found}

Initialize population $\mathcal{P} = \{x_1, \dots, x_N\}$ with random chromosomes of length $L$\;
Evaluate fitness $f(x_i)$ for all $x_i \in \mathcal{P}$\;

\For{$G$}{
    Select random sub-population $\mathcal{S} \subset \mathcal{P}$ of size $k$\;
    Sort $\mathcal{S}$ greatest to least with regards to fitness\;
    
    Perform single-point crossover on $\mathcal{S}_{(1)}$, $\mathcal{S}_{(2)}$ to generate offspring $c_1, c_2$\;

    \For{$i \in \{1, 2\}$}{
        Apply Gaussian mutation to $c_i$ with rate $\mu$ and scale $\sigma$\;
        Evaluate and record fitness $f(c_i)$\;
        Replace $S_{(-i)}$ with $c_i$ in $\mathcal{P}$\;
    }

    Update best individual if needed\;
}
\Return best individual found\;
\end{algorithm}

\subsection{Training Loop}
\label{subsec:training}

The training loop integrates SCOPE with an SSGA in a simulation environment, detailed in Section~\ref{sec:environment}, to evolve and evaluate gait parameters. At the beginning of each generation, the SSGA creates an offspring as a candidate solution. This solution includes both the weights and the biases for the evolved policy that controls the hexapod's gait. During evaluation, SCOPE applies the 2D DCT and applies evolved weight vectors as detailed in Section~\ref{subsec:SCOPE}. The resulting gait parameters control the phase, amplitude, and offset of the CPG controlling each motor.

Each candidate is evaluated in a 15-second episode, divided into 5 sub-episodes of 3 seconds. At the beginning of each sub-episode, the controller collects the most recent frames, feeds them through the SCOPE and uses the output as the new gait parameters. These parameters are used throughout the sub-episode. This cycle repeats for each sub-episode during the full episode. After the full 15-second episode, the robot computes its performance by calculating the Euclidean distance difference between its locations at the start and beginning of the episode, and this is used by SSGA as a fitness value. The overall training loop with different main components is visualized in Figure~\ref{fig:main}.

This loop of trial and feedback is done continuously throughout the simulation. After the initialization of the random population, the robot is not reset between evaluations. Individuals are exchanged inside the controller without pausing or resetting the simulation.

\section{Environment}
\label{sec:environment}

\begin{figure}[t]
    \centering
    \includegraphics[width=0.5\linewidth]{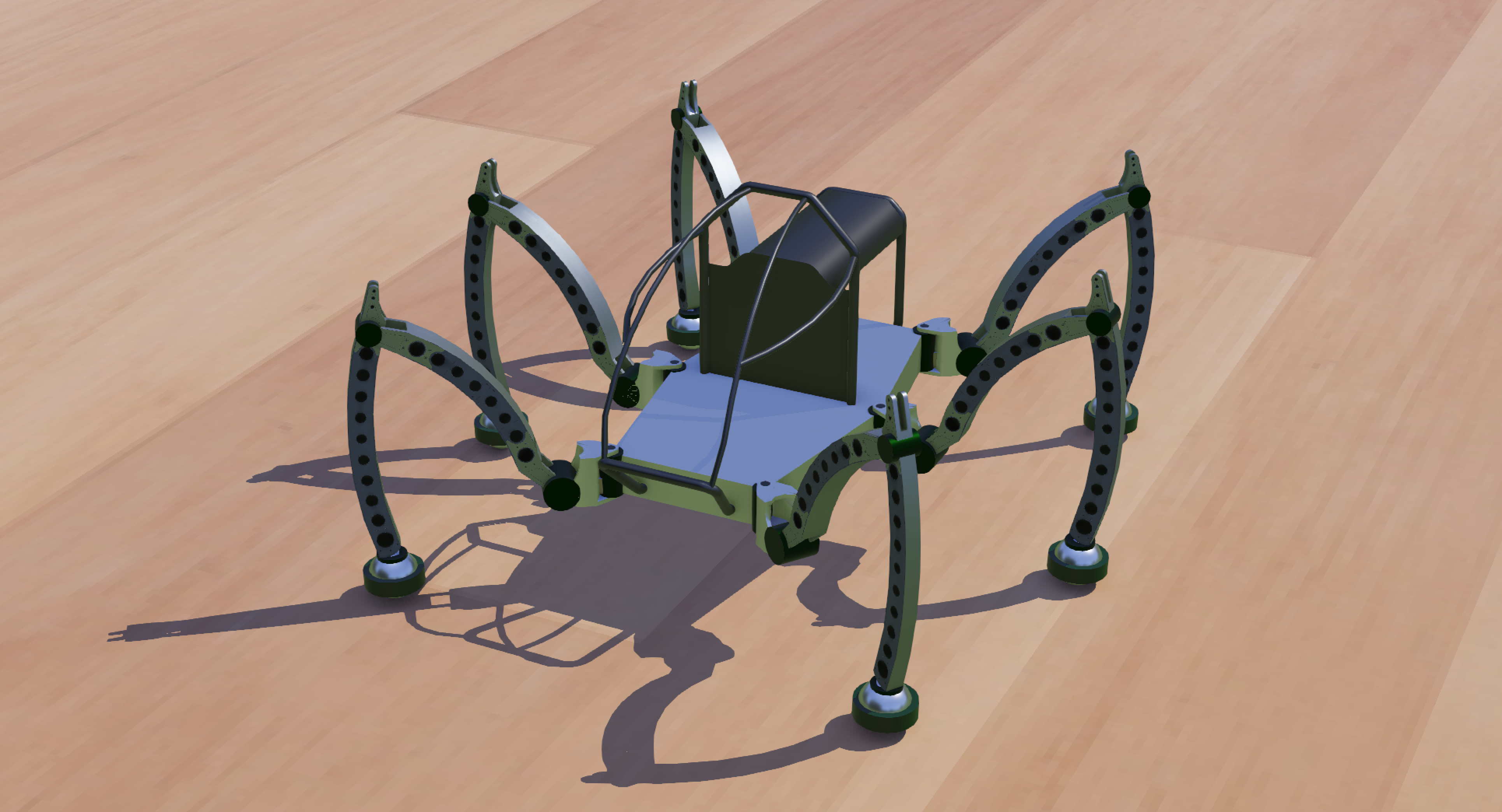}
    \caption{A photo of the simulation environment showing the simulated version of the Mantis robot and the training arena.}
    \label{fig:sim}
\end{figure}

All experiments are conducted in WeBots, an open source robotics simulator \cite{WEBOTS}. The simulated robot is inspired by the Mantis hexapod robot designed and built by Micromagic Systems. The Mantis robot has six legs, each with coxa, femur, and tibia joints, resulting in 18 independent motors. The simulation environment, shown in Figure~\ref{fig:sim}, is a simple flat surface with standard physics settings, with all evaluations carried out in identical conditions. 

Each motor operates using positional control and is limited to reasonable angular limits. The coxa joints are limited between $-15^\circ$ and $15^\circ$, the femur joints between $40^\circ$ and $80^\circ$, and the Tibia joints between $-150^\circ$ and $-105^\circ$. These joint limits are enforced throughout the simulation to prevent unrealistic movement and to avoid leg collisions that could result in tangling behavior.

The simulation runs with a frame length of 32 milliseconds. Each motor includes a position sensor, which provides the current position of the motor at each frame. The velocity and acceleration values are calculated with the position readings. For each motor, a state vector $\mathbf{m}_i = [p_i, v_i, a_i]$ is formed, where $p_i$, $v_i$, and $a_i$ are the position, velocity, and acceleration of the $i$-th motor, respectively. These values are stacked to form our initial $18 \times 3$ matrix:
\[
\mathbf{M}_{\text{raw}} =
\begin{bmatrix}
\mathbf{m}_1 \\
\mathbf{m}_2 \\
\mathbf{m}_3 \\
\mathbf{m}_4 \\
\vdots \\
\mathbf{m}_{18}
\end{bmatrix}
=
\begin{bmatrix}
p_1 & v_1 & a_1 \\
p_2 & v_2 & a_2 \\
p_3 & v_3 & a_3 \\
p_4 & v_4 & a_4 \\
\vdots & \vdots & \vdots \\
p_{18} & v_{18} & a_{18}
\end{bmatrix}
\in \mathbb{R}^{18 \times 3}.
\]

This matrix is restructured into a $6 \times 9$ matrix $\mathbf{M} \in \mathbb{R}^{6 \times 9}$, where each row corresponds to one leg:
\[
\mathbf{M} =
\begin{bmatrix}
\mathbf{m}_1 & \mathbf{m}_2 & \mathbf{m}_3 \\
\mathbf{m}_4 & \mathbf{m}_5 & \mathbf{m}_6 \\
\vdots & \vdots & \vdots \\
\mathbf{m}_{16} & \mathbf{m}_{17} & \mathbf{m}_{18}
\end{bmatrix}
=
\begin{bmatrix}
p_1 & v_1 & a_1 & p_2 & v_2 & a_2 & p_3 & v_3 & a_3 \\
p_4 & v_4 & a_4 & p_5 & v_5 & a_5 & p_6 & v_6 & a_6 \\
\vdots & \vdots & \vdots & \vdots & \vdots & \vdots & \vdots & \vdots & \vdots \\
p_{16} & v_{16} & a_{16} & p_{17} & v_{17} & a_{17} & p_{18} & v_{18} & a_{18}
\end{bmatrix}
\in \mathbb{R}^{6 \times 9}
\]
This transformation reflects the structure of the robot, as it represents each leg as a single row. We then construct the full input matrix by concatenating the most recent 50 robot states horizontally, forming a matrix of shape \( \mathbb{R}^{6 \times 450} \). Each state is a matrix \( \mathbf{M}_{(t)} \in \mathbb{R}^{6 \times 9} \), and we define the full input matrix \( \hat{\mathbf{M}} \) as:
\[
\hat{\mathbf{M}} =
\begin{bmatrix}
\mathbf{M}_{(t)} & \mathbf{M}_{(t-1)} & \mathbf{M}_{(t-2)} & \cdots & \mathbf{M}_{(t-49)}
\end{bmatrix}
\in \mathbb{R}^{6 \times 450}
\]
This input matrix $\hat{\mathbf{M}}$ is used by the Sparse Cosine Optimized Policy Evolution, detailed in Section~\ref{subsec:SCOPE}.

\section{Results}
\label{sec:Results}

To evaluate the performance implications of the Sparse Cosine Optimized Policy Optimization (SCOPE), we compared the performance of a baseline evolutionary algorithm to an evolutionary algorithm augmented by SCOPE. Our core evolutionary algorithm is a Steady-State Genetic Algorithm, with the addition of SCOPE being the only change between the baseline and augmented version.

The baseline version evolves a direct transformation of the raw time-series sensor data to gait parameters without any dimensionality reduction. A pose matrix of shape $6 \times 9$ is used to form a time-series of the previous 50 poses, producing a matrix of shape $6\times 450$ with $2{,}700$ total inputs. Each individual solution contains a weight and bias vector, each with a length equal to the total number of inputs, resulting in a total of $5{,}400$ parameters per solution. While this approach allows the algorithm to capture the patterns in the data, the large parameter space limits the efficiency of the search, making it harder to converge toward an optimal solution. 

When we augment the evolutionary algorithm with SCOPE, the discrete cosine transformation is applied to the same $6 \times 450$ input, reducing input to a compressed representation of $6 \times 9$ features. This leads to a large reduction in the number of parameters per individual solution, with each individual evolving only 54 weights and 54 biases, for a total of 108 parameters. This 98\% reduction in parameter count significantly improves search efficiency.

\begin{figure*}[t]
    \centering
    \includegraphics[width=1\linewidth]{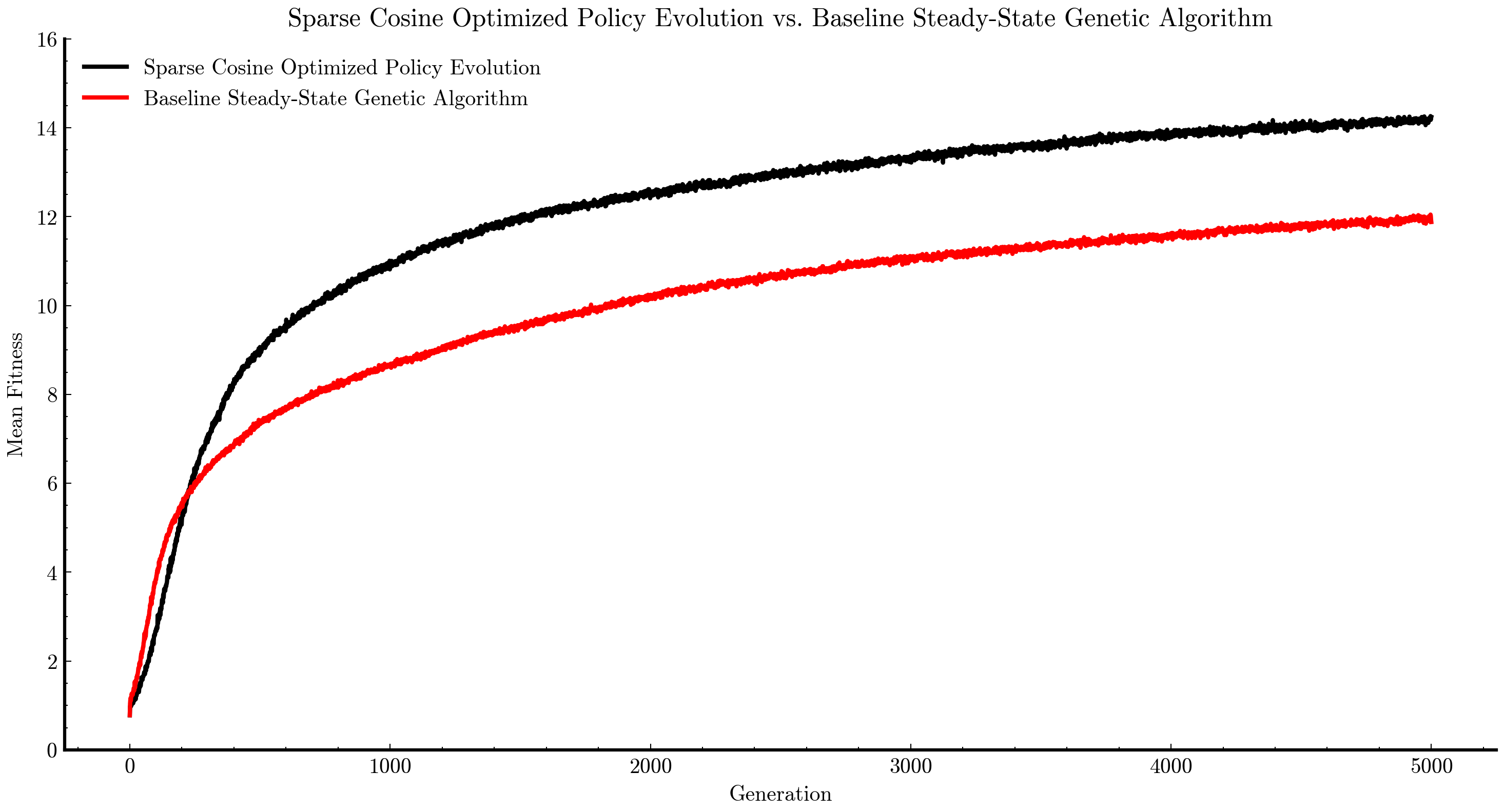}
    \caption{This graph shows the mean fitness curve of a steady state genetic algorithm optimizing the weights of the gait generation policy with and without the use of a discrete cosine transformation. Each algorithm was allowed to evolve over 5000 generations, and made use of identical mutation rates, crossover methods, and population sizes. To address the inherent randomness of these algorithms, the graph shows an average of 500 individual trials for both algorithms, with $1{,}000$ trials in total.}
    \label{fig:5k-fitness}
\end{figure*}

Each method was trained independently over $5{,}000$ generations using the training procedure described in Sections~\ref{subsec:training} and~\ref{sec:environment}, over 500 independent trials. The fitness metric throughout these trials was the Euclidean distance traveled by the robot throughout the 15-second episode. Figure~\ref{fig:5k-fitness} shows the mean fitness curves of both approaches over $5{,}000$ generations.

The trials that included SCOPE augmentation, and therefore fewer free parameters, consistently outperformed the baseline method. On average, the addition of SCOPE resulted in a higher fitness values with a consistent performance gap present throughout the full length of the trial. Augmenting the learning algorithm with SCOPE resulted in an average fitness of $14.242$ while the baseline had an average fitness of $11.880$, showing a 20\% increase on average. Statistical testing confirms the significance of this result, a one-sided Mann-Whitney U test results in a U-Statistic of $176{,}524.00$ and a $p$-value of $7.998 \times 10^{-30}$.

\section{Conclusion}
\label{sec:conclusion}

This work introduced Sparse Cosine Optimized Policy Evolution (SCOPE), a novel method for compressing high-dimensional input data using the Discrete Cosine Transform (DCT) in order to improve the sample efficiency and tractability of evolutionary algorithms. By transforming the raw sensor time-series data into the frequency domain and retaining only the lowest-frequency components, SCOPE preserves task-relevant structure while discarding redundancy and fine-grained variation.

We demonstrated the integration of SCOPE with a Steady-State Genetic Algorithm in the domain of hexapod gait learning. SCOPE compresses the pose data from the robot’s previous 50 poses, reducing a 2,700-dimensional signal to just 54 values. This compressed signal is used by an evolved policy to decide the gait parameters of the robot. Despite this aggressive compression, the resulting controller achieves superior performance compared to a baseline model that does not perform compression, attaining a 20\% improvement in average fitness across 500 independent training runs.

SCOPE achieves these gains without introducing any domain-specific assumptions about the robot or its environment. This makes it broadly applicable to other learning problems in robotics and control where input dimensionality poses a bottleneck.

\section{Future Work}
\label{futurework}

While this study focused on the application of SCOPE to hexapod gait generation, the underlying method is broadly applicable to a wide range of control and decision-making problems. In ongoing work, we are extending SCOPE to domains such as game-playing, where input observations tend to be high-dimensional, noisy, and partially redundant. These are conditions under which the Discrete Cosine Transform (DCT) is especially effective.

Unlike the structured and noise-free sensor input of our hexapod robot, environments like the Atari Learning Environment or other visual-based benchmarks introduce substantial irrelevant or distracting information \cite{ale}. In these settings, many observed features (e.g., static background pixels or repetitive UI elements) do not contribute meaningfully to the decision-making process. We hypothesize that the frequency-based compression offered by SCOPE can improve both learning efficiency and policy robustness by filtering out noise and highlighting the dominant temporal or spatial features.

Although we refrained from applying additional sparsity to the DCT coefficients in this study, due to the precision and relevance of the robotic input data, we are investigating the role of coefficient sparsification in noisy domains. In particular, we aim to evaluate whether removing the lowest percentile of DCT coefficients improves generalization and reduces overfitting in settings where inputs are stochastic or contain substantial irrelevant information.

Additionally, we envision integrating SCOPE with other evolutionary strategies, such as Covariance Matrix Adaptation (CMA-ES) or Quality Diversity (QD) algorithms, to test whether its compression benefits carry over across different optimization regimes.

\bibliographystyle{splncs04}
\bibliography{references}

\end{document}